\documentclass[journal]{IEEEtran}

\usepackage{cite}

\ifCLASSINFOpdf
 \usepackage[pdftex]{graphicx}
 \usepackage{subfigure}
\else
\fi

\usepackage{amsmath}
\usepackage{array}
\usepackage{booktabs}
\usepackage{multirow}
\usepackage{graphicx}
\usepackage[normalem]{ulem}
\useunder{\uline}{\ul}{}

\begin{document}

\title{An Automated Machine Learning (AutoML) Method for Driving Distraction Detection Based on Lane-Keeping Performance}

\author{Chen~Chai,~\IEEEmembership{Member,~IEEE,}~Juanwu~Lu,~Xuan~Jiang,\\Xiupeng~Shi,~Zeng~Zeng,~\IEEEmembership{Senior~Member,~IEEE}%

\thanks{This study was jointly sponsored by the National Key Research and Development Program of China (project number: 2018YFB1600502), and the Chinese National Science Foundation (61803283).}%

\thanks{Chen Chai and Juanwu Lu are with the College of Transportation Engineering, Tongji University, Shanghai, 201804, China. Chen Chai is also with the Key Laboratory of Road and Trafﬁc Engineering of the Ministry of Education, Tongji University, Shanghai, 201804, China. (e-mail: chaichen@tongji.edu.cn; lujuanwu@gmail.com).}%

\thanks{Xuan Jiang is with the Department of Civil and Environmental Engineering, University of California, Berkeley, CA, 94720, United States. (e-mail: j503440616@berkeley.edu).}%

\thanks{Xiupeng Shi and Zeng Zeng are with Institute for Infocomm Research, Agency for Science Technology and Research (A*STAR), 138632, Singapore (e-mail: shix@i2r.a-star.edu.sg; zengz@i2r.a-star.edu.sg).}%

\thanks{Corresponding authors: Zeng Zeng.}
}




\maketitle

\begin{abstract}

With the enrichment of smartphones, driving distractions caused by phone usages have become a threat to driving safety. A promising way to mitigate driving distractions is to detect them and give real-time safety warnings. However, existing detection algorithms face two major challenges, low user acceptance caused by in-vehicle camera sensors, and uncertain accuracy of pre-trained models due to drivers’ individual differences. Therefore, this study proposes a domain-specific automated machine learning (AutoML) to self-learn the optimal models to detect distraction based on lane-keeping performance data. The AutoML integrates the key modeling steps into an auto-optimizable pipeline, including knowledge-based feature extraction, feature selection by recursive feature elimination (RFE), algorithm selection, and hyperparameter auto-tuning by Bayesian optimization. An AutoML method based on XGBoost, termed AutoGBM, is built as the classifier for prediction and feature ranking. The model is tested based on driving simulator experiments of three driving distractions caused by phone usage: browsing short messages, browsing long messages, and answering a phone call. The proposed AutoGBM method is found to be reliable and promising to predict phone-related driving distractions, which achieves satisfactory results prediction, with a predictive power of 80\% on group level and 90\% on individual level accuracy. Moreover, the results also evoke the fact that each distraction types and drivers require different optimized hyperparameters values, which reconfirm the necessity of utilizing AutoML to detect driving distractions. The purposed AutoGBM not only produces better performance with fewer features; but also provides data-driven insights about system design.

\end{abstract}

\begin{IEEEkeywords}
Automated Machine Learning, Driving Distraction Detection, Lane Keeping Performance, Driver Activity Recognition.
\end{IEEEkeywords}

\IEEEpeerreviewmaketitle

\section{Introduction}
\IEEEPARstart {D}{riving} distraction is a global threat to road safety. According to the National Highway Traffic Safety Administration presents, 8\% of the traffic fatal cases are caused by driving distraction and leads to 2,841 deaths in 2018\cite{dot812926}. Similar results can be found in SafetyNet Accident Causation Database that 32\% of the crashes involved at least one driver, rider, or pedestrian who was described as ‘Inattentive’ or ‘Distracted’\cite{talbot2013exploring}. Several causations are responsible for these emerging phenomena, including the increasing phone usage for calling, texting and web browsing during driving. Emerging fatal cases calls for attention from the public on detecting and preventing driving distraction.

Several studies have presented the impacts of driving distraction and how in-vehicle supporting systems could reduce drivers’ risk of accidents\cite{liu2006vehicle,liao2015impact, green2004driver}. Recently, one of the main focuses on driving distraction detection has been the implementation of machine learning for detecting distraction behavior. Existing studies utilize two major categories of data inputs, namely physiological indicators and driving performance. The former includes measurements of driving conditions, e.g., body movements\cite{halverson2012classifying,jung2014driver,liu2015driver,wang2017binocular,jiang2018safedrive}, organ activities\cite{rahman2016case}, or combined\cite{miyaji2008driver}. Nevertheless, most physiological indicators require specific equipment (e.g., electrocardiogram machines) to be on body for tracking, which is not straightforward to deploy and may cause additional distractions.

As a promising alternative, driving performance is recently adopted to assess distraction, with advantages such as minimum influences on the drivers, being easy to deploy, massive data collection from both naturalistic and simulated driving environment \cite{li2017visual, zhang2017identification}. Notwithstanding, three major limitations exist challenging driving distraction detection. First, professional knowledge and massive efforts are needed for model building, including prior knowledge to extraction high-quality features, massive trials to find the optimal machine learning solutions. Second, individual deviations are important but are not well studied. Individual deviations may lower the robustness and accuracy of the proposed algorithm, thus induce difficulties for training. Last but not the least, the imbalance of classes within the training dataset (e.g., rare occurrences of distracted driving) could cause biases towards the safe driving samples during modeling and evaluation. There is a perennial quest to develop an automated and reliable method to perform optimized modeling based on given data and application scenarios.

The main contributions of the work can be summarized as follows:

(1) We introduce an automated machine learning (AutoML) method to self-learn the optimal models based on given data for driving distraction detection based on lane-keeping performance.

(2) We explore a set of solutions to address the main above-mentioned challenges, such as AutoML considering domain-specific prediction tasks, knowledge-guided feature extraction, and feature ranking and selection.

(3) We diagnose the heterogeneity of driving distraction detection, as well as individual deviations, and delineate a data-driven AutoML system to detect distraction in a non-invasive manner for smart driving, which extends the scope of solutions as relying on pre-trained models.

This study presents an AutoML for driving distraction detection based on vehicle movement data. Section 2 reviews the literature. Section 3 elaborates the methodology of AutoML. Section 4 presents the analysis of experiment results and data-driven insights. The final two sections cover discussion and conclusions.

\section{Literature Review}

\subsection{Driving Distraction}
Driver distraction is defined as “the diversion of attention away from activities critical for safe driving towards a competing activity”\cite{regan2008driver}. Hence, data that reflect the driver’s condition is essential for detecting such distractions. Initial research on driving distraction detection could be dated back to the 1990s, where measurable decrements with physiological signs and changes in the driving performance became two of the major classes of inputs utilized for the detection\cite{knipling1994vehicle}.

Physiological indicators are long used as dominant signals to reveal the status of a driver. Body movements (e.g., eye movements, head rotations) are widely used to build the linkage with driving distraction behavior \cite{liu2006vehicle,liu2015driver,liao2015impact,wang2017binocular}. Based on eye movement data collected from ten participants ranging in age from 19 to 26 years, pupil size and percentage of eye closure are useful for predicting workload \cite{halverson2012classifying}. A monitoring system with embedded electrocardiogram sensors on the steering wheel is developed with viability tested with two participants \cite{jung2014driver}. However, dynamically tracking such physiological signs requires specific devices (e.g., ECG sensors), which are disturbing and hard to set up or manage. Besides, an intrusive data collecting procedure makes it not practical in daily use\cite{hansen2017driver,jiang2019denoising}.

Therefore, driving performance indicators as an alternative are widely adopted in recent studies, which are relatively easy to extract from trajectory data of either a simulated or a naturalistic environment. Typical indicators include speed, acceleration, and distance headway. For example, a hybrid of Genetic Algorithm and SVM is developed to effectively determine 3 different driver distraction states, based on inputs including the mean and standard deviation (std) of speed, mean car-following distance, the std values of steering wheel angle, side deviation, longitudinal and lateral accelerations \cite{zhang2017identification}. Based on similar input features, promising results are retrieved via testing by naturalistic driving data of 108 randomly sampled passenger-car drivers \cite{li2017visual}. In addition to a single type of data input, driving performance indicators are combined with physiological signals such as eye movements, to further improve the overall performance \cite{ma2016distraction}.

In short, existing models for driving distraction detection mainly make use of physiological signals and driving performance indicators as inputs for training. The latter has significant advantages in terms of easy to obtain and less intrusive. Promising detection results could be obtained with state-of-the-art algorithms. Nevertheless, three major challenges remain under-explored:

\subsubsection{Costs in training} Conventional procedures of machine learning include a plethora of design decisions, such as architecture search and parameter tuning, which creates barriers for field application.

\subsubsection{Individual differences} Distraction features vary among individuals and outliers within the dataset could lower the performance of proposed algorithms.

\subsubsection{Imbalanced Classes} Distracted driving rarely occurs in daily scenarios, which leads to a significant disparity in the number of samples of distracted driving versus safe driving. It further creates difficulties for modeling and parameter tuning.

\subsection{AutoML}
AutoML is a process of learning to learn, which assembles necessary machine learning steps into an end-to-end pipeline and automates the pipeline to get the optimal features, algorithms, and hyperparameter settings that return the best performance \cite{shi2019feature}. Building an efficient and interpretable AutoML system is inherently challenging. Achieving an outperformed performance depends not only on the fundamental power of the core algorithms, but also on an optimal-configured end-to-end pipeline, including careful data processing, high-quality feature engineering, sophisticated hyperparameter tuning, among others \cite{feurer2019auto, eggensperger2019pitfalls}. Moreover, AutoML for domain-specific tasks faces extra technical and practical challenges.

First, feature engineering is an important aspect to improve the predictive ability as well as interpretability, which is the process of extracting information from data based on domain knowledge or algorithms (e.g., neural networks), as the input of machine learning algorithms \cite{garcia2016big}.

Second, the performance improvement of a given algorithm is also leveraged on well-tuning of hyper-parameter settings, which usually entails trying out the best one from all possible values manually \cite{feurer2019auto}. Generally, hyperparameter optimization (HPO), meta-learning, and neural architecture search (NAS) are three typical AutoML implementations \cite{feurer2019hyperparameter}. Thereinto, Bayesian Optimization is a powerful method for black-box function optimization \cite{bergstra2013making, mendoza2016towards}, and has obtained satisfying state-of-the-art results recently in hyperparameters tuning of deep neural networks for natural language processing \cite{melis2018state}, image classification \cite{snoek2012practical,snoek2015scalable}, and speech recognition \cite{dahl2013improving}. 

Third, considering the time and computation constraints in real-time applications, it is more practical to have a lean and fast-response AutoML system that can achieve high performance with limited data samples and computing costs \cite{shahriari2015taking}. Thereinto, decision-tree based ensemble learning algorithms (e.g., XGBoost, lightGBM) have demonstrated advantages in both performance and interpretability \cite{chen2016xgboost}.

Certainly, domain-specific AutoML enables a lot of benefits but relevant research is still much lacking.

\section{Driving Distraction Experiments}

To acquire the dataset for training, we conducted a driver distraction experiment based on a driving simulator located in Tongji University to observe and collect driving behavioral data.

\subsection{Participants}
A total of 28 drivers (18 males and 10 females) aged from 21 to 48 (mean=26.25, std=6.58) were recruited. All of these participants had valid driving licenses. Herein, most of them had driving  experience for more than 3 years (mean=3.46, std=6.58) and all of them once drove on a mountain freeway. We offered a cash reimbursement for each participant.

\subsection{Apparatus}
The driving simulator experiment was conducted using Tongji University's high-fidelity driving simulator, which was the most advanced in China. Fig. 1 demonstrates the layout of our apparatus, a Renault Megane III passenger car was placed in a dome with 5 projectors. These projectors help to provide a $250^{\circ}\times 40^{\circ}$ field of view with up to $1000\times 1050$ high resolution. The dome was attached to an 8 degree-of-freedom motion system with an X-Y range of $20m\times 5m$, which was capable of producing actual sounds such as noises of vehicle engines and real senses of acceleration, deceleration, braking, and yawing. In combined with SCANeR studio software for generating the scenario and display, the simulator could give an immersive experience to the drivers.

\begin{figure}
  \centering
  \includegraphics[width=\linewidth]{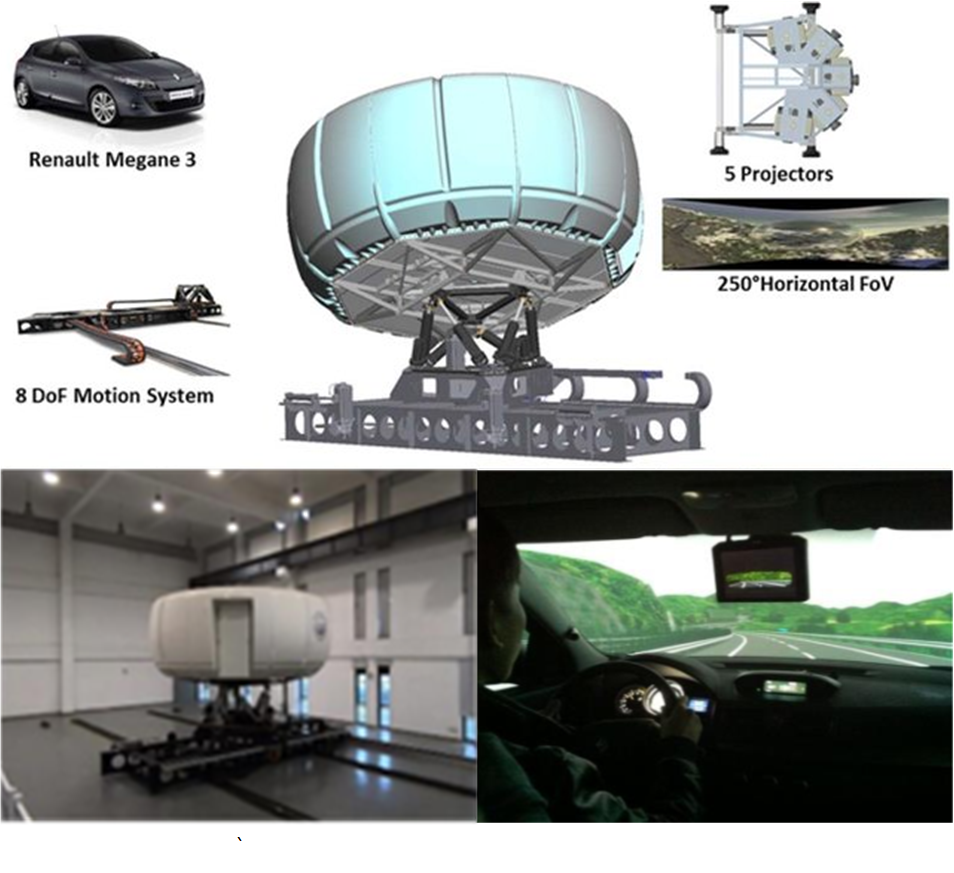}
  \caption{Tongji University driving simulator}
  \label{Fig: Simulator}
\end{figure}

\subsection{Procedure and Tasks}
A two-way four-lane mountain freeway (see Fig. 2) with a length of approximately 24km was generated for our experiment to reduce the complexity and variety of the scene and deviation of results caused by drivers’ exploration. Vegetative covers were also introduced to increase the sense of reality, but no other vehicles were placed in the scenario for reducing environmental distraction and controlling variables. Data were retrieved from the simulator with a sampling rate of 20 Hz.

\begin{figure}
  \centering
  \includegraphics[width=\linewidth]{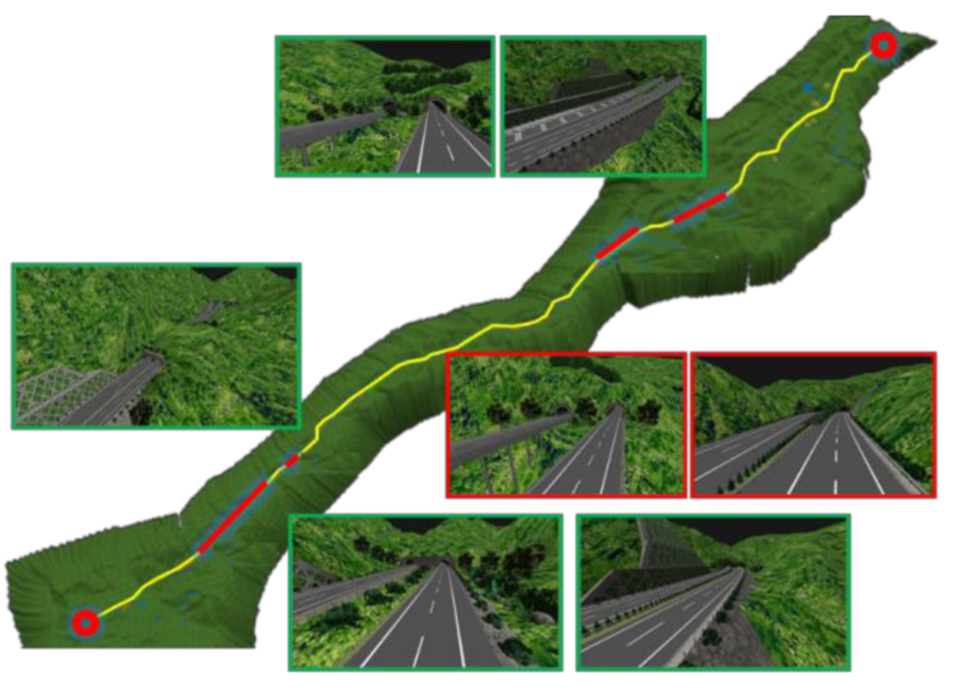}
  \caption{Simulation scenario}
  \label{Fig: scenario}
\end{figure}

The procedure of this experiment included two separate drives: a baseline drive without any subtask, and a distracted drive where drivers were asked to undertake a variety of phone use subtasks (browsing short messages, answering phone calls, browsing long messages). Drivers were first required to take a practice drive and informed about the requirements of these subtasks. Each drive lasted around 11 minutes with a recommended speed of 60 km/h. As demonstrated by Fig. 3, the following three subtasks were assigned during driving by a preset procedure:

\begin{figure}
  \centering
  \includegraphics[width=\linewidth]{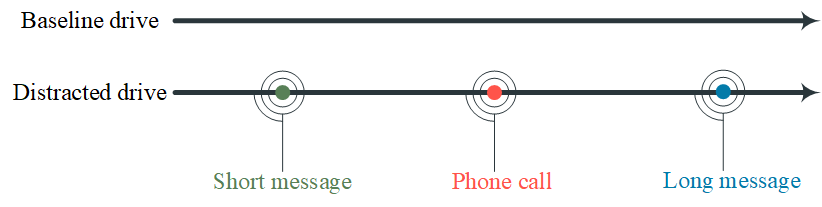}
  \caption{The participants were required to undertake no secondary task in a baseline drive, and several subtasks in a distracted drive for comparison.}
  \label{Fig: procedure}
\end{figure}

\subsubsection{Short message} Drivers were asked to read and understand a short message - \textit{The meeting tomorrow has been postponed until 11 a.m.}

\subsubsection{Phone call} Drivers would receive a phone call with questions (e.g., \textit{Do you have a class tomorrow morning?}) and be required to deliver proper answers.

\subsubsection{Long message} Drivers were asked to read and understand a long message - \textit{To popularize the knowledge of emergency care at the site of the accident and improve the ability of self-help and mutual rescue, a cardiopulmonary resuscitation (CPR) first-aid training is now held in the college of architecture and urban planning room B5 at 8:30, September $21^{st}$. Participants who pass the examination after the training will receive the lifeguard certificate. Please attend on time.}

\subsection{Questionnaires}

Aiming at revealing the individual differences in driving behavior patterns of participants, driving behavior evaluation was conducted using a modified Driving Behavior Scale with 33 featured undesirable driving behaviors. Drivers were required to rate their likelihood of exhibiting these behaviors on a scale of 1 (Never) to 5 (Nearly all the time). The questions were then grouped into 5 categories, including transient rule violations, fixed rule violations, misjudgment, risky driving exposure, and driver mood. Transient rule violations indicate risky driving behaviors that can happen multiple times in one journey; fixed rule violations, on the contrary, are stable across the journey. Misjudgment unravels driving errors on traffic conditions of speeds and space headways; risky exposure means the exposure of risky driving behaviors; driver mood measures emotion status and managements. The results of each category score are shown in TABLE \ref{tab:qresults}.

\begin{table}[htbp]
  \caption{Means and standard deviations of the participants' scores}
  \label{tab:qresults}
  \centering
  \begin{tabular}{p{13em} p{5em} p{5em}}
   \toprule
   \multicolumn{1}{c}{\multirow{2}{*}{\textbf{Items}}} & \multicolumn{2}{c}{\textbf{Scores}}   \\
   \cmidrule(l){2-3} 
   \multicolumn{1}{c}{}   & Mean & Std. \\
   \midrule
   Transient rule violations (TR)   & 2.35   & 0.60   \\
   Fixed rule violations (FI)   & 1.59   & 0.41   \\
   Misjudgment (MS)   & 1.78   & 0.50   \\
   Risky driving exposure (EX)   & 2.81   & 0.54   \\
   Driver mood (DM)   & 2.37   & 0.74   \\   
   \bottomrule
   \end{tabular}%
\end{table}

\subsection{Measures of Lane-keeping Performance}

\begin{table*}[t]
  \caption{Metrics and Features}
  \centering
  \label{tab:features}
  \begin{tabular}{@{}lll@{}}
    \toprule
    \textbf{Metric} & \textbf{Formula} & \textbf{Feature} \\
    \midrule
    Mean ($M$) & $M=\frac{1}{N}\Sigma_{i=1}^{N}x_{i}$ & $LV_{M}$, $LA_{M}$, $YV_{M}$, $YA_{M}$, $LD_{M}$, $LDL_{M}$, $LDR_{M}$ \\ 
    Standard Deviation ($SD$) & $SD=\sqrt{\frac{1}{N-1}\Sigma_{i=1}^N\left(x_i-M\right)^2}$ & $LV_{SD}$, $LA_{SD}$, $YV_{SD}$, $YA_{SD}$, $LD_{SD}$, $LDL_{SD}$, $LDR_{SD}$\\ 
    Range ($R$) & $R=x_{max}-x_{min}$ & $LD_{R}$ \\
    Coefficient of Variation ($C_v$) & $C_v=\frac{SD}{M}$ & $LDL_{C_{v}}$, $LDR_{C_{v}}$ \\ 
    Quartile Coefficient of Variation ($Q_{cv}$) & $Q_{cv}=\frac{x_{75\%}-x_{25\%}}{x_{75\%}+x_{25\%}}$ & $LDL_{Q_{cv}}$, $LDR_{Q_{cv}}$ \\
    \bottomrule
  \end{tabular}%
\end{table*}

In this experiment, we extracted 7 preliminary features that could reflect the driver’s lane-keeping performance, namely, lateral velocity (V), lateral acceleration (LA), yaw angular velocity (YV), yaw angular acceleration (YA), as well as the lane departures to the centerline (LD), to the left side (LDL), and to the right side (LDR), respectively. Thereinto, yaw angle refers to the deviation angle between driving direction and lane centerline, and lane departures refer to the distance between the centerline of the vehicle and the corresponding roadsides or lane centerline.

The limited dimensionality of raw data input, however, was not effective enough for our algorithm. Hence, we further implemented time aggregation with a granularity of 1.0 second to generate features that represented volatility, such as range, mean, and standard deviation. Besides, measures like the coefficient of variation ($C_v$)\cite{wali2018driving} and quartile coefficient of variation ($Q_{cv}$)\cite{bonett2006confidence} from previous studies were also used to further describe fluctuations of lane departures. Variables and their corresponding metrics are as described in TABLE~\ref{tab:features}. Finally, a total of 19 features generated with regard to seven lane-keeping performance indicators were used in our algorithm.

\section{Methodology}

\subsection{Automated Driving Distraction Detection}

\begin{figure}[htbp]
  \centering
  \includegraphics[width=\linewidth]{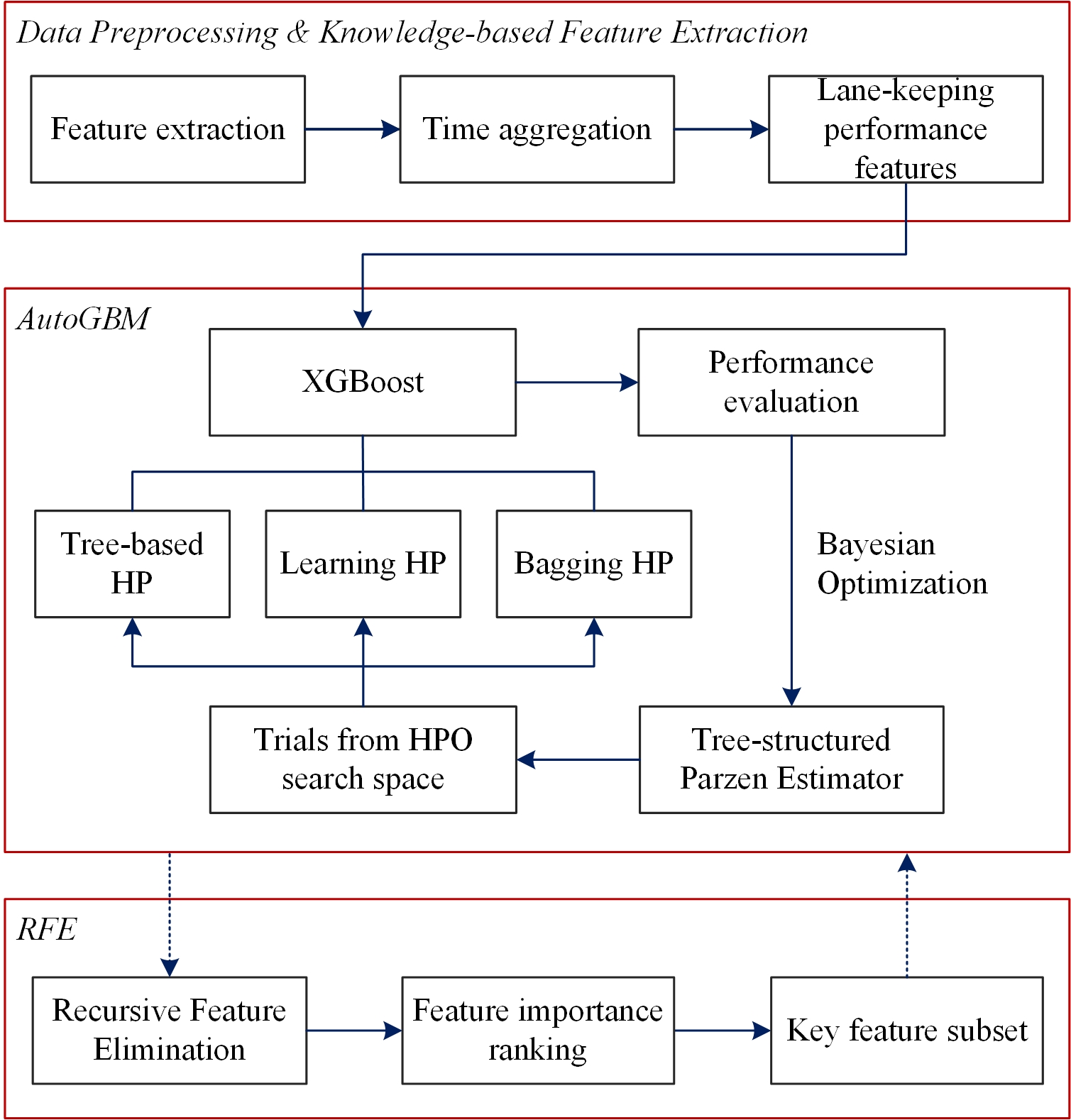}
  \caption{Automated Driving Distraction Detection.}
  \label{Fig: flowchart}
\end{figure}

A task-specific AutoML framework is designed to achieve self-optimized modeling for given data. To address the main challenges in terms of high costs of modeling, individual differences, and imbalanced classes, a data-driven AutoML is proposed to calibrate an optimal combination of hyperparameters for driving distraction detection based on lane-keeping performance, as depicted in Fig.~\ref{Fig: flowchart}. Typical modules of the AutoML pipeline include massive feature extraction based on domain knowledge (e.g., lane-keeping performance features), model selection and hyperparameter auto-tuning by Bayesian optimization, as well as learning-based feature selection. 

Herein, an auto-optimizable XGBoost model (termed AutoGBM) is built as the key classifier throughout the pipeline, and the key features are identified based on recursive feature elimination (RFE). Bayesian optimization guides the auto-tuning process to find the best-suited models and hyperparameter values within specific computation capacity and time constraints. Besides, the pipeline also incorporates pre-processing steps such as time aggregation, imbalanced data resampling, noise filtering, among others.

\subsection{AutoGBM}

AutoML is optimized by finding the best algorithm settings and hyperparameter values that maximize the model performance on validation sets. For hyperparameter optimization, Tree-structured Parzen Estimator (TPE) algorithm is implemented and will be described in the next sub-section.

The AutoGBM implemented the auto-tuning of XGBoost or other tree-based gradient boosting models. XGBoost is used as the basic classifier of AutoML considering the advantages in terms of regularization, effective tree pruning, and high performance. Previous studies have proved that XGBoost has better performance than most existing driving distraction detection  methods, such as support vector machine and random forest. The objective function of XGBoost consists of two parts, namely, a loss to measure the difference between the predicted score and the real score, and regularization terms, which is expressed as:

$$O b j=\sum_{i=1}^{n} l\left(y_{i}, \hat{y}_{i}\right)+\sum_{k=1}^{K} \Omega\left(f_{k}\right)$$

where $l$ is the training loss function, and $\Omega$ is the regularization term. The training loss measures how predictive the model is on training data. 

This auto-tuning process usually requires trying out all promising settings and values, which entails huge combinations. Besides, the mapping from the hyperparameters to the performance is known as a black-box function, which is tedious and expensive to optimize.

\subsection{Hyperparameter Auto-tuning}

\begin{table*}[!t]
  \caption{Hyperparameters Configuration Space}
  \label{tab:hyperparams}
  \resizebox{\textwidth}{!}{%
  \begin{tabular}{@{}cccl@{}}
    \toprule
    \textbf{Hyperparameter}   & \textbf{Distribution} & \textbf{Range} & \multicolumn{1}{c}{\textbf{Description}} \\ \midrule
    Number of Estimators & Uniform & [30, 150] & Total number of trees generated in XGBoost model.\\
    Learning Rate & Log-uniform & $\left[ln\left(0.05\right),ln\left(0.3\right)\right]$ & Shrinkage factor for updates in each boosting step to prevent overfitting. \\
    Level Subsample Ratio & Uniform & [0.6, 1.0] & Subsample ratio of columns for each new depth level. \\
    Tree Subsample Ratio & Uniform & [0.6, 1.0] & Subsample ratio of columns for constructing each tree. \\
    Instances Subsample Ratio & Uniform & [0.6, 1.0] & Subsample ratio of the training samples. \\
    Maximum Depth & Stochastic & \{3, 4, 5, 6\} & The maximum depth of a tree. \\
    Minimum Child Weight & Uniform & [0.6, 1.0]  & The minimum sum of instance weight necessary in a child. \\
    $\alpha$ & Uniform & [0.0, 1.0] & L1 regularization term on weights. \\
    $\gamma$ & Uniform & [0.0, 1.0] & The minimum loss value reduction necessary to make a further partition on a leaf node of the tree. \\
    $\lambda$ & Uniform & [0.4, 1.0] & L2 regularization term on weights. \\ \bottomrule
  \end{tabular}%
  }
\end{table*}

\subsubsection{TPE-based Bayesian Optimisation}~

To deliver the optimized performance, model selection and hyperparameter tuning are conducted based on Bayesian optimization. The framework is also scalable for a series of algorithms, which have the model selection and tuning of specific hyperparameter settings. This process usually requires trying out all promising algorithms and hyperparameter values $M(a,h)$ to find the ones that maximize the model performance \cite{feurer2019hyperparameter}, which entails massive combinations. 

Bayesian optimization is efficient for the tedious black-box tuning. It constructs a probabilistic surrogate model $p(y\mid x)$ of the objective function $f(x)$ that maps input values $x\in M(a,h)$ to a probability of a loss, making it easier to optimize than the actual $f(x)$ \cite{shahriari2015taking}. Besides, by reasoning from past search results, the next trials can concentrate on more promising ones, which reduces the number of trials while finding a good optimum. The Bayesian-based auto-tuning of model selection and hyperparameter values is represented as:

\begin{equation*}
    x^{*} =\arg \max _{x \in M(a, h)} f(x),
\end{equation*}
\begin{equation*}
\begin{split}
      EI_{y^{*}}(x) &=\int_{-\infty}^{y^{*}}\left(y^{*}-y\right) p(y|x) d y \\ 
      &\propto \left(\gamma+\frac{g(x)}{l(x)}(1-\gamma)\right)^{-1},
\end{split}
\end{equation*}
\begin{equation*}
    \gamma =p\left(y<y^{*}\right),
\end{equation*}
\begin{equation*}
    p(x|y) =\left\{\begin{array}{ll}
    l(x) & \text {if~} y<y^{*} \\
    g(x) & \text {if~} y \geq y^{*}
    \end{array}\right..
\end{equation*}

Herein, $p(x|y)$ and quantile $\gamma$ are built based on Tree-structured Parzen Estimator (TPE) to produce a predictive posterior distribution of models such as XGBoost $M(a,h)$ over the performance of past results and form two non-parametric densities $l(x)$ and $g(x)$, which then guide the exploration of the model domain space \cite{snoek2015scalable, bergstra2015hyperopt}. The models $x^{*}$ with the highest expected improvement $EI_{y^{*}}(x)$ are selected for the next trials, which are expected to potentially minimize the loss function, namely, increase the performance.

\subsubsection{Hyperparameter Search Space}~

Complex and high-dimensional search space of the hyperparameters has been one of the several challenges which make hyperparameter optimization a hard problem in practice \cite{feurer2019hyperparameter}. However, a previous study demonstrated empirically that only some hyperparameters, in most cases, have significant impacts on overall performance \cite{bergstra2012random}. To reduce the complexity and accelerate the training process, we define the configuration space for our algorithm with nested stochastic function expressions. Herein, a uniform distribution is used to describe the Number of Estimators, which is the total amount of trees generated for the XGBoost model. The learning rate is described using a logarithmic uniform distribution, which has been informed by prior practice \cite{surmenok2017estimating}. Other hyperparameters and their corresponding configuration space are demonstrated in TABLE~\ref{tab:hyperparams}.

\subsubsection{Objective Functions}~

The objective function is estimated by a set of metrics, scored on testing datasets via stratified cross-validation. For imbalanced classification such as distraction detection, the recall, precision and AUPRC (area under the precision-recall curve) are recommended to evaluate model performance \cite{shi2019feature, lever2016classification}. Additional metrics include accuracy and AUC (area under the receiver operating characteristic curve). The metrics are calculated for each class, and macro-averaged mean (over all classes) and weighted mean (based on the number of true instances for each class) values are calculated, respectively \cite{lever2016classification}.

\subsection{Feature Ranking by RFE}

AutoML relies on two key aspects to improve performance from input perspectives, namely, knowledge-guided feature extraction, and identification of important features. Massive features are extracted to provide the fundamental of predictive ability. A learning-based feature selection approach is developed to identify the most important features. The interpretations of feature extraction and related terminologies are summarized in TABLE~\ref{tab:features}.

The feature selection procedure ranks and filters a set of relatively important features based on AutoGBM, and then permutes to find an optimal feature combination using recursive feature elimination (RFE) \cite{shi2020automated}.

To make the selection from massive features more efficient, the feature relative importance can also be measured by the split weight and average gain of each feature, which are generated in the process of XGBoost fitting. The feature filtering by importance ranking can reduce the feature space efficiently, and RFE further considers the interactions among features, and identifies the best combination.

\section{Analysis}

Six optimization experiments are conducted utilizing different datasets. The first experiment used data collected from a total of 28 drivers without any separation of different types of distraction. The $2^{nd}$ to $4^{th}$ experiments were developed for different distraction tasks, namely, short message, long message, and phone call. The $5^{th}$ and $6^{th}$ experiments were developed on individual drivers. Two typical types of drivers are selected according to the driving behavior questionnaire, namely, a young male driver with a high-risk driving profile, and an old male driver with a low-risk driving profile. Results of the experiments are discussed in the following session, including the AutoGBM procedures, modeling performance, feature analysis, and data-driven insights.

\subsection{Auto-tuning by Bayesian Optimization}

AutoGBM delivered optimized models for driving distraction detection based on the hyperparameter auto-tuning of XGBoost by Bayesian optimization. The hyperparameters and domain distributions are listed in TABLE \ref{tab:hyperparams}. The domain space to search hyperparameters is created as centered around the pre-tested values and is then refined in subsequent searches \cite{shi2020automated}.

\begin{figure}[!t]
\flushleft
\includegraphics[width=3.4in]{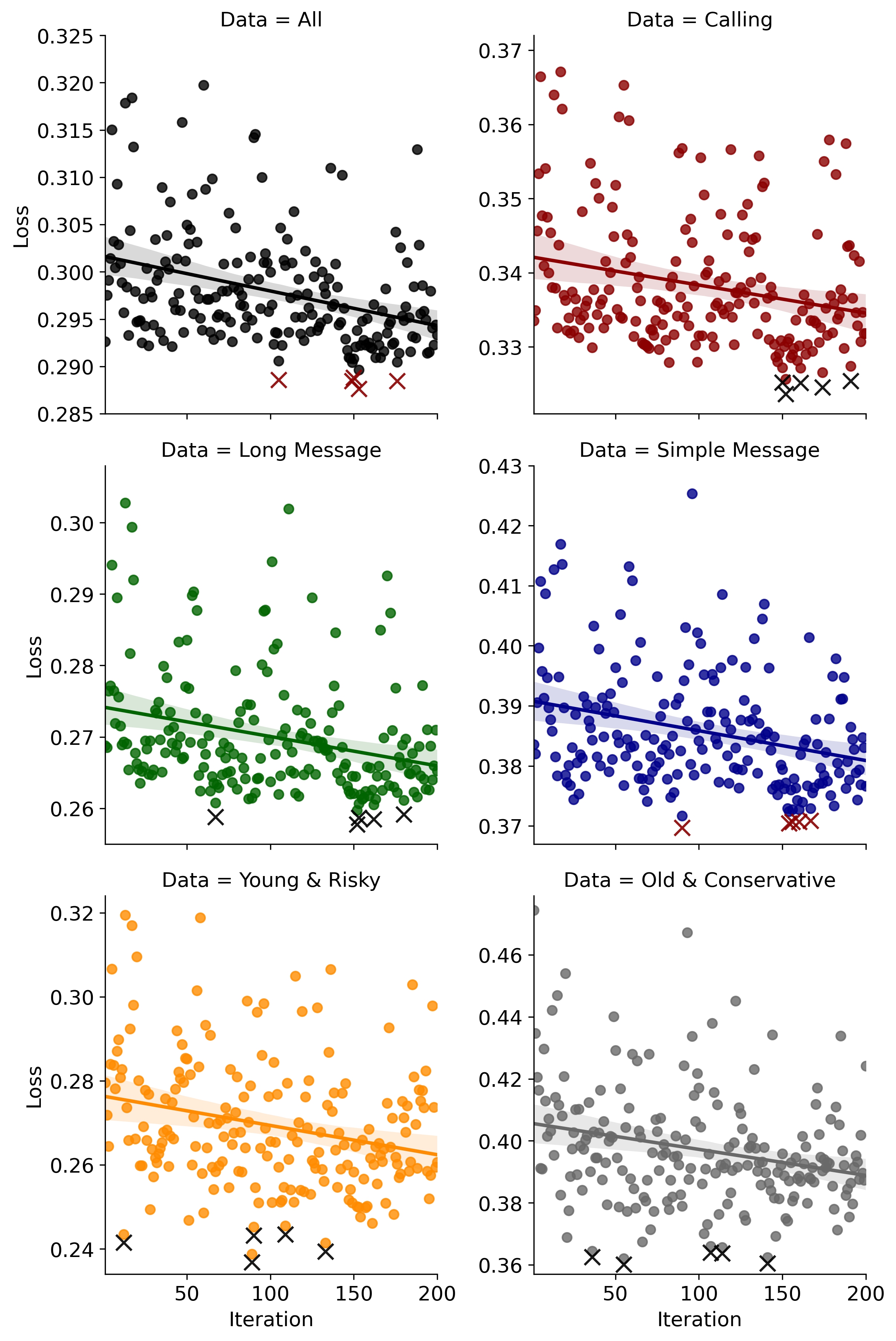}
\caption{Bayesian-based model auto-tuning. Loss versus the iterations is plotted to inspect the auto-modeling process. The overall performance is improved (i.e., more tries of lower loss over time as expected), indicating that AutoGBM is searching for better hyperparameters based on given data.}
\label{fig_AutoML}
\end{figure}

The basis of AutoGBM is configured by the ensemble of boosted decision trees. To improve performance, tree hyperparameters (e.g., maximum tree depth, splitting weight) are tuned to control model complexity, and the learning rate is tuned to shrink the boosting process, which makes fitting more conservative. The logarithmic uniform distribution is used for the learning rate because it varies across several orders of magnitude. Besides, bagging hyperparameters (e.g., instance subsampling, feature subsampling) reduce the variance by decorrelation, which helps to improve the model robustness against noise. The sampling of values in the domain is equally likely (uniform). The total number of estimators is set to 300, and early stopping is also applied to stop the training when validation scores have not improved for several iterations (e.g., herein 10\% of total estimators).

\begin{figure*}[t]
\flushleft
\includegraphics[width=6.8in]{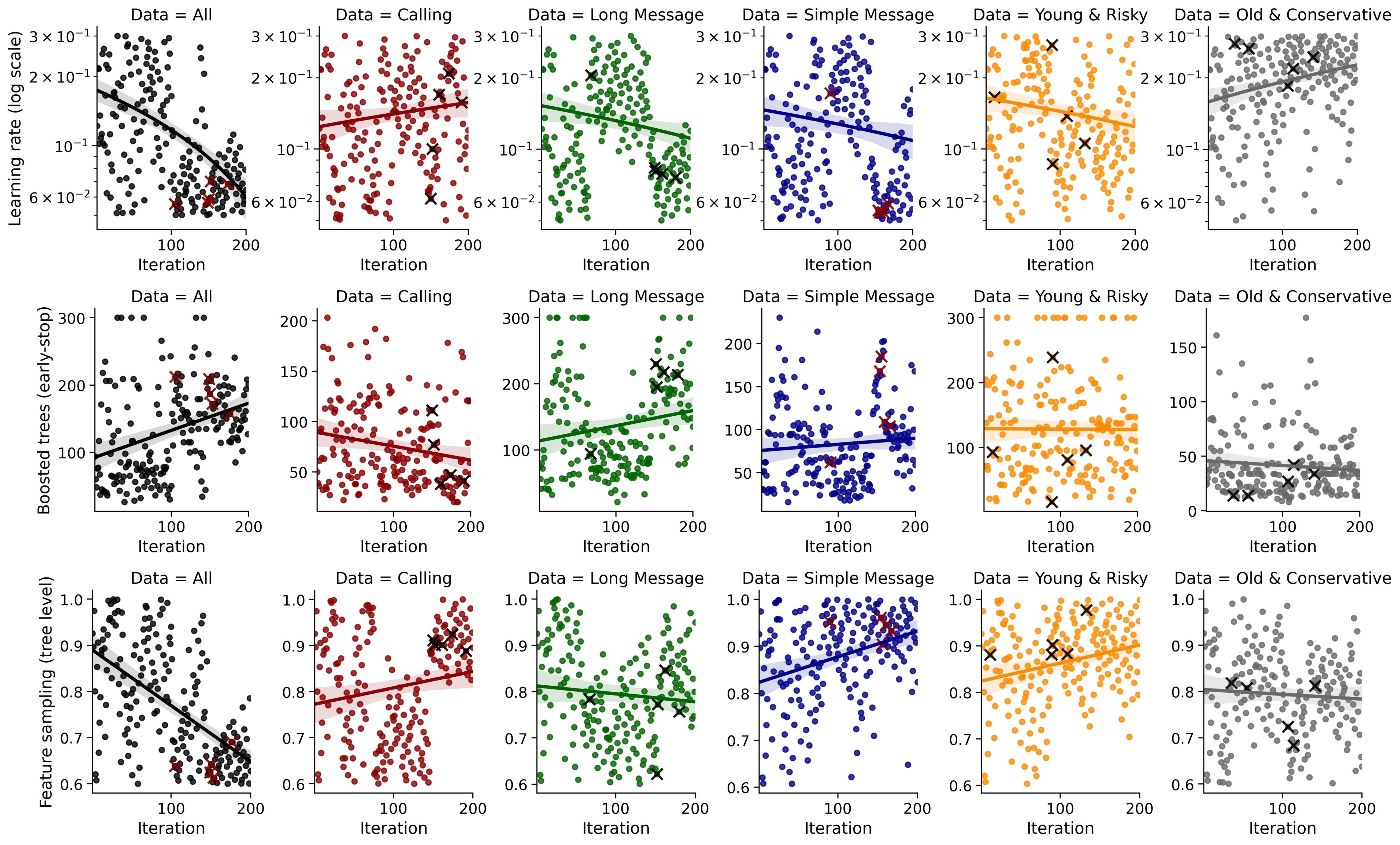}
\caption{Hyperparameters versus the iterations plotted to inspect the searching for better hyperparameter values. Trends show that the model focuses more trials on more promising values.}
\label{fig_iter}
\end{figure*}

AutoGBM screens out better hyperparameter values from the domain space based on TPE, which evaluates the relationship between previous loss and hyperparameter settings, and chooses next trials of values that are more promising to deliver better performance. Herein, the objective function is to minimize the log loss via 10-fold stratified cross-validation, and 200 iterations of trials are used.

The performance and hyperparameters versus the iterations are plotted to inspect the auto-tuning process, as shown in Fig.~\ref{fig_AutoML} and Fig.~\ref{fig_iter}, respectively. The dark x-markers indicate the top 5 optimal results. The average performance increase over time (conversely the loss decreases) as expected, indicating AutoGBM is trying better hyperparameter values. As the search progresses, the auto-tuning switches between exploration (e.g., trying new values) and exploitation (e.g., selecting values within the ranges of better past results), which is more efficient compared to uninformed random or grid search methods.

\begin{figure*}[t]
\flushleft
\includegraphics[width=6.8in]{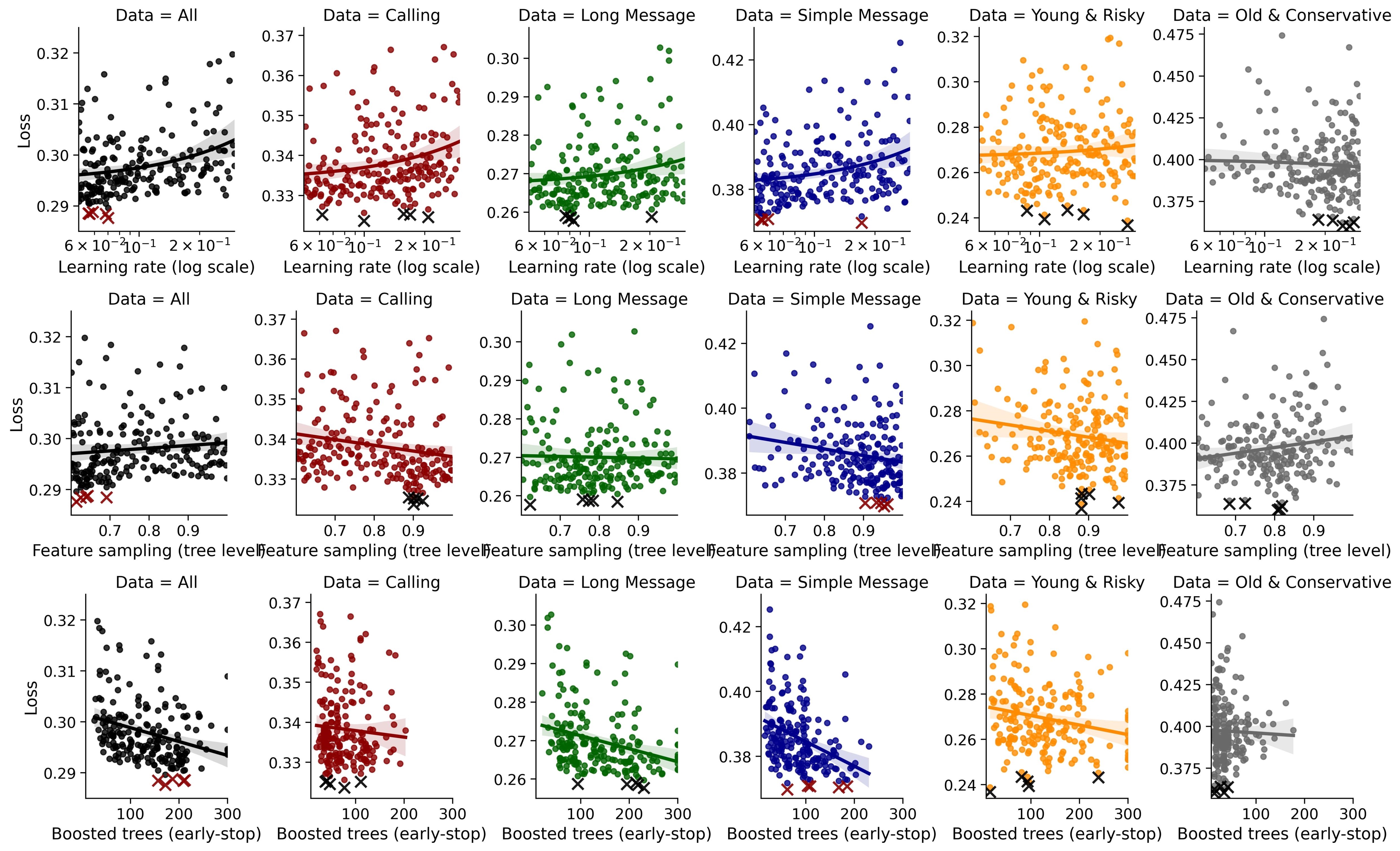}
\caption{Distribution of hyperparameters values with better performance.}
\label{fig_loss}
\end{figure*}

The AutoGBM also reveals key hyperparameters with greater impacts to improve detection performance. The relationships between loss and key hyperparameters are plotted in Fig.~\ref{fig_loss}, including learning rates, feature subsample ratios, numbers of estimators by early stop. Trends showed that the Bayesian optimization tends to concentrate (i.e., place more probability) the search on evaluating more promising values. The identification of key hyperparameters also contributes to the better design of domain-specific AutoML.

\subsection{Modeling Performance}

\begin{table*}[t]
\centering
\caption{Optimized Hyper-parameters and Performance Scoring}
\label{tab:harresults}
\begin{tabular}{p{15em} p{6em} p{6em} p{6em} p{6em} p{6em} p{6em} }
\toprule
   & \multirow{2}{*}{\textbf{ALL}} & \multicolumn{3}{c}{\textbf{Task-based split}}   & \multicolumn{2}{c}{\textbf{Driver-based split}}   \\ 
   \cmidrule(l){3-5} 
   \cmidrule(l){6-7} 
   &   & Long Message  & Simple Message   & Calling   & Young~\&~Risky   & Old~\&~Conservative  \\ \midrule
\textbf{I. Hyperparameters} & \multicolumn{1}{l}{\textbf{}} & \multicolumn{1}{l}{\textbf{}} & \multicolumn{1}{l}{\textbf{}} & \multicolumn{1}{l}{\textbf{}} & \multicolumn{1}{l}{\textbf{}} & \multicolumn{1}{l}{\textbf{}} \\
Number of Estimators   & 155   & 128   & 81   & 47   & 16   & 34   \\
Learning Rate   & 0.083   & 0.114   & 0.110   & 0.209   & 0.275   & 0.206   \\
Level Subsample Ratio   & 0.810   & 0.601   & 0.955   & 0.774   & 0.643   & 0.870   \\
Tree Subsample Ratio   & 0.618   & 0.723   & 0.786   & 0.925   & 0.881   & 0.899   \\
Instances Subsample Ratio   & 0.810   & 0.884   & 0.706   & 0.971   & 0.967   & 0.418   \\
Maximum Depth   & 4   & 5   & 3   & 6   & 6   & 3   \\
Minimum Child Weight   & 0.729   & 0.703   & 0.774   & 0.963   & 0.970   & 0.974   \\
$\alpha$   & 0.078   & 0.828   & 0.053   & 0.351   & 0.092   & 0.145   \\
$\lambda$   & 0.271   & 0.627   & 0.871   & 0.314   & 0.225   & 0.292   \\
$\gamma$   & 0.909   & 0.691   & 0.433   & 0.651   & 0.443   & 0.134   \\
\textbf{II. Performance}   & \textbf{}   & \textbf{}   & \textbf{}   & \textbf{}   & \textbf{}   & \textbf{}   \\
Loss   & 0.295   & 0.267   & 0.385   & 0.327   & 0.239   & 0.378   \\
Accuracy   & 0.830   & 0.877   & 0.794   & 0.845   & 0.905   & 0.857   \\
\bottomrule
\end{tabular}%
\end{table*}

\begin{figure}[!t]
\flushleft
\includegraphics[width=3.0in]{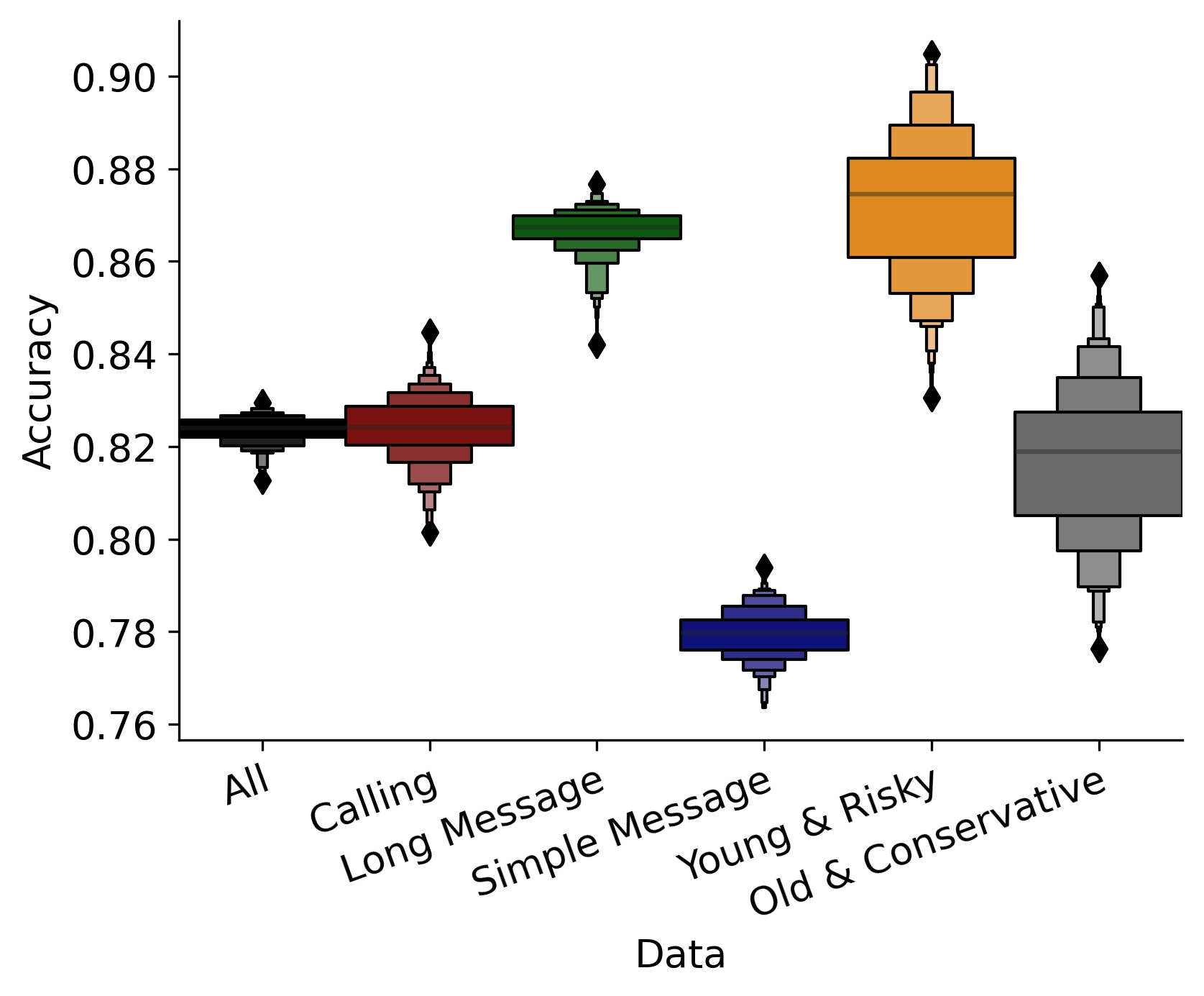}
\caption{Heterogeneity analysis of prediction performance based on AutoGBM.}
\label{fig_Acc}
\end{figure}

After Bayesian-based auto-tuning, AutoGBM constructs an ensemble of several versatile models, by adding the model that maximizes ensemble validation performance through iteration. The automatic ensemble is more robust and less prone to over-fit, compared in favor of selecting one best model. The optimized hyperparameter values and corresponding performance scores are listed in TABLE \ref{tab:harresults}.

Different experiments present diverse characteristics according to the performance improvement by hyperparameter auto-tuning. The heterogeneity is summarised by a letter-value plot in Fig.~\ref{fig_Acc}. Letter-value plots display further letter values than Boxplots (i.e., display the median and quartiles only), which include more detailed information about the tails behavior and afford more precise estimates of corresponding quantiles, thus reflects more details on performance comparison. 

As shown in Fig.~\ref{fig_Acc}, the AutoGBM model achieves higher accuracy when detecting driving distraction caused by a complex task like reading a long message, and shows promising results when applied to young and risky drivers.

Furthermore, according to optimized hyperparameters shown in TABLE \ref{tab:harresults}, AutoML method revealed differences between different driving distraction tasks and individual drivers. At the task level, the most apparent difference occurred in the optimized number of estimators. The reading message, especially long message, required more estimators than detecting a phone call. It requires more estimators to detect the distraction of the old and conservative driver at the individual level. This is consistent with existing driving distraction studies, which have found most experienced drivers compensate lane keeping performance by reducing speed and paying more attention to vehicle performance while using phones.


\subsection{Feature Ranking and Selection}

\begin{figure}[!t]
\flushleft
\includegraphics[width=3.4in]{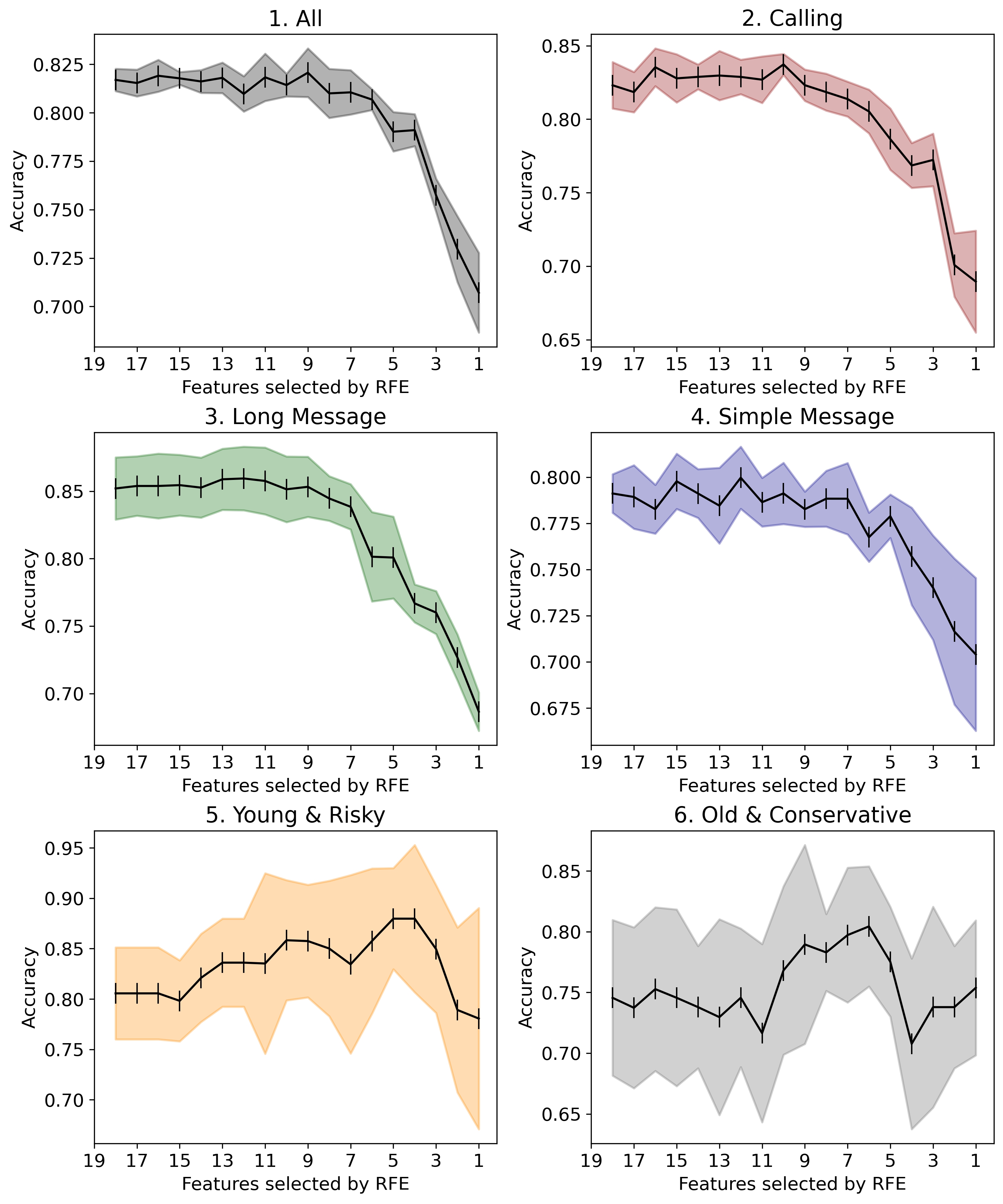}
\caption{Feature self-selection by RFE and learning performance.}
\label{fig_RFE}
\end{figure}

\begin{table*}[t]
\centering
\caption{Important feature ranking for different tasks and drivers}
\label{tab:feature_ranking}
\begin{tabular}{p{7em} p{7em} p{7em} p{7em} p{7em} p{7em} p{7em} }
\bottomrule
    \multicolumn{1}{c}{\multirow{2}{*}{\textbf{Feature ranking}}} & \multicolumn{1}{c}{\multirow{2}{*}{\textbf{All}}} & \multicolumn{3}{c}{\textbf{Task-based split}}    & \multicolumn{2}{c}{\textbf{Driver-based split}}    \\
    \cmidrule(l){3-5}
    \cmidrule(l){6-7}
\multicolumn{1}{l}{}    & \multicolumn{1}{c}{}    & \multicolumn{1}{c}{Long Message} & \multicolumn{1}{c}{Simple Message} & \multicolumn{1}{c}{Calling} & \multicolumn{1}{c}{Young~\&~Risky} & \multicolumn{1}{c}{Old~\&~Conservative} \\ \hline
1   & SDLDR   & MeanLDL   & SDLDL   & SDYawv   & MeanLDR   & QCVLDR   \\
2   & SDLDL   & MeanYawA    & QCVLDL    & MeanYawA   & QCVLDL    & CVLDR    \\
3   & SDLPA   & QCVLDR    & MeanYawA    & MeanLDL    & AmpLP   & AmpLP    \\
4   & MeanYawA    & QCVLDL    & CVLDR   & QCVLDL   & SDLDL   & MeanLDR    \\
5   & MeanLDR   & SDLDL   & MeanLDR   & MeanLPA    & MeanYawv    & Meanv    \\
6   & CVLDR   & SDLDR   & QCVLDR    & SDLDL    & MeanLPA   & MeanLP   \\
7   & SDYawv    & CVLDR   & SDYawv    & QCVLDR   & CVLDR   & MeanLPA    \\
8   & QCVLDR    & MeanLDR   & MeanLPA   & SDLPA    & QCVLDR    & CVLDL    \\
9   & QCVLDL    & Meanv   & MeanLDL   & CVLDR    & SDLP    & MeanYawv   \\
10    & Meanv   & SDYawv    & Meanv   & MeanYawv   & SDLDR   & SDLP   \\
11    & MeanLDL   & MeanYawv    & MeanYawv    & SDLDR    & MeanLDL   & SDv    \\
12    & MeanYawv    & CVLDL   & SDLDR   & CVLDL    & CVLDL   & SDLDL    \\
13    & MeanLPA   & MeanLPA   & SDYawA    & Meanv    & Meanv   & SDLDR    \\
14    & CVLDL   & AmpLP   & AmpLP   & SDYawA   & SDv   & MeanLDL    \\
15    & SDLP    & MeanLP    & CVLDL   & MeanLDR    & MeanYawA    & QCVLDL   \\
16    & SDYawA    & SDLPA   & MeanLP    & AmpLP    & MeanLP    & SDYawv   \\
17    & MeanLP    & SDLP    & SDLP    & MeanLP   & SDYawv    & SDYawA   \\
18    & AmpLP   & SDv   & SDv   & SDLP   & SDLPA   & SDLPA    \\
19    & SDv   & SDYawA    & SDLPA   & SDv    & SDYawA    & MeanYawA   \\ 
\bottomrule
\end{tabular}
\end{table*}

The identification of important features provides data-driven insights on data processing and sensor fusion and guides the direction of deep-level feature extraction. Besides, the identification of optimal feature subsets contributes to improving performance and reducing computation cost. Learning-based feature selection is an iterative process in terms of both modeling and RFE aspects. The RFE iteration is conducted by recursively pruning the features with the least permutation importance from the current set. With the selected optimal feature subset, the AutoGBM could be updated alternately. Herein, we configure the classifier used for RFE based on the auto-tuned hyperparameters, except that feature subsampling hyperparameters are set to 1.0 to avoided randomness in subsampling.

The RFE-based feature selection and corresponding learning performance are illustrated in Fig.~\ref{fig_RFE}. The learning performance is measured by the mean and standard deviation values of cross-validation accuracy, which are represented by the line charts and shadows, respectively. The key features are selected based on the trade-off of less complexity, better performance, as well as lower variance. The selected features are listed in TABLE~\ref{tab:feature_ranking}.

The feature selection based on RFE is to interact among features by re-training. Then the importance of each feature and optimal feature combination subsets will be identified. Besides, in the training process, the split weight and average gain for each feature in tree building are generated, which can be normalized to calculate the weight-based and gain-based relative importance scores, respectively. Weight-based selection generally favors features with more classes, and gain-based selection is biased towards the ones with stronger signals (e.g., Gini impurity) \cite{shi2020automated}. Thus, RFE is more reliable to assess and identify the feature importance. Moreover, rooted in decision trees, AutoGBM with knowledge-specific features has the advantages of high transparency, being robust to noise and randomness, as well as better tolerance to missing data, etc.

\section{Discussion}
\subsection{Limitations and Future Work}

In this paper, we propose an AutoGBM model with a promising performance in addressing individual differences (e.g., young and risky drivers) and imbalanced classes. It helps improve road safety by enabling an automated and predictive method to identify early risk signs of driving distraction. However, the model is not fully prepared for predictions concerning accuracy on some specific types of subtasks, including reading a short message. One potential way for improvement is to enlarge the amount of data specifically on these subtasks, which require further data acquisition and preprocessing.

For the sake of improving modeling, in-depth mining of viable lane-keeping performance features and a larger scale of high-quality data sources are suggested, which contribute to revealing a wider range of potential mutual relationships. From the perspective of methodology, introducing new algorithms can help refine the overall performance.

\subsection{Application Potentials}

Driving distraction detection with automated machine learning is an emerging area with huge potentials in improving in-vehicle safety assistant systems. There are multiple folds for the application potentials. From the perspective of the industry, an automated and proactive solution can reduce costs in system development and algorithm optimization, which can accelerate the research and development footprints. From the perspective of drivers, an advanced system could help guarantee safety on roads, by risk warnings and driving recommendations.

\section{Conclusions}

This study developed an AutoGBM approach to detect driving distraction based on lane-keeping performance. Compared with existing studies, the end-to-end AutoML is effective and flexible to predict both reading messages and answering phone calls. Utilizing lane-keeping performance instead of in-vehicle camera sensors also improves user security and acceptance. In summary, this research has contributed to the domain knowledge in four areas and provide data-driven insights about driver monitoring system design, as highlighted in the following.

(1) The AutoGBM method enables the auto-optimization of driving distraction detection pipeline adapted onto data, which integrates the main steps including, lane-keeping feature extraction, hyperparameter tuning by Bayesian optimization, and feature ranking based on RFE. Based on Bayesian optimization, the algorithm selection and hyperparameter tuning are self-learned, and the best models with the minimum loss are generated. Besides, AutoGBM is rooted in decision trees, interpretable prediction rules can thus be generated.

(2) The AutoGBM method achieves satisfactory results of driving distraction detection, which has a predictive power of 80-90\% overall accuracy of 28 drivers, and greater than 90\% accuracy at the individual level. Based on the unified AutoML procedure embedded, the detection performances of various phone usage tasks have been compared, which provides data-driven insights about developing driver monitoring system. Moreover, the AutoML reveals the most important knowledge-based features for assessing driver’s individual differences, which uncovers useful insights about adoptive in-vehicle AI functions. At the event level, detecting complex message reading is found to require more estimators, deeper model, larger L1 and L2 regularization term on weights to achieve higher accuracy. At the individual level, detect the distraction of young and risky driving drivers is found to be easier than the older and conservative driver, with fewer estimators and higher accuracy.

(3) Although further evaluations of a larger variety of drivers and distractions should be conducted to evaluate and improve the developed model, the results show to enrich the potential of applications towards the intelligent vehicle. With the power of intrinsic and high-accuracy driver state monitoring, the developed methods are useful towards a range of advanced solutions such as detection of aggressive driving, driving fatigue, and driving anger.


\ifCLASSOPTIONcaptionsoff
  \newpage
\fi

\bibliographystyle{IEEEtran}
\bibliography{X}

\vfill

\vfill





\vfill


\end{document}